\documentclass{article}
\usepackage{spconf,amsmath,graphicx}
\usepackage{subcaption}
\usepackage{booktabs}
\usepackage[hyphens]{url}
\usepackage{comment}
\usepackage{here}

\title{Background Mixup Data Augmentation for \\ Hand and Object-in-Contact Detection}
\name{Koya Tango, Takehiko Ohkawa, Ryosuke Furuta, Yoichi Sato
\thanks{This work was supported by JST AIP Acceleration Research Grant Number JPMJCR20U1 and JST ACT-X Grant Number JPMJAX2007.}}
\address{The University of Tokyo, Tokyo, Japan}

\usepackage{fancyhdr}
\usepackage{lipsum}

\fancypagestyle{specialfooter}{%
  \fancyhf{}
  
  \fancyfoot[C]{\footnotesize © 2022 IEEE. Personal use of this material is permitted. Permission from IEEE must be obtained for all other uses, in any current or future media, including reprinting/republishing this material for advertising or promotional purposes, creating new collective works, for resale or redistribution to servers or lists, or reuse of any copyrighted component of this work in other works.}
}

\begin{document}
%
\thispagestyle{specialfooter}

\maketitle
\begin{abstract}
Detecting the positions of human hands and objects-in-contact (hand-object detection) in each video frame is vital for understanding human activities from videos.
For training an object detector, a method called Mixup, which overlays two training images to mitigate data bias, has been empirically shown to be effective for data augmentation.
However, in hand-object detection, mixing two hand-manipulation images produces unintended biases, e.g., the concentration of hands and objects in a specific region degrades the ability of the hand-object detector to identify object boundaries.
We propose a data-augmentation method called Background Mixup that leverages data-mixing regularization while reducing the unintended effects in hand-object detection.
Instead of mixing two images where a hand and an object in contact appear, we mix a target training image with background images without hands and objects-in-contact extracted from external image sources, and use the mixed images for training the detector.
Our experiments demonstrated that the proposed method can effectively reduce false positives and improve the performance of hand-object detection in both supervised and semi-supervised learning settings.

\end{abstract}
\section{Introduction}
\label{sec:intro}

Detecting the positions of a person's hands and an object-in-contact (hand-object detection) from an image provides an important clue for understanding how the person interacts with the physical world. This hand-object detection is applicable to recognizing a person's primitive actions, such as ``taking" or ``pushing", and logging the person's activity of interacting with the environment~\cite{yagi2021go}. Shan et al.~\cite{Shan20} built a hand-object detector for localizing hands and interacting objects on a large-scale dataset collected in naturalistic house-holding situations, such as in kitchen~\cite{damen2018scaling, li2018eye, sigurdsson2018actor}, DIY~\cite{Shan20}, and craft work~\cite{Shan20, sigurdsson2018actor}.

However, a hand-object detector trained on such house-holding images may not be well generalized to other hand-manipulation images. For instance, the images in biological laboratories or factories have significantly different data distribution from the daily scenes used in training. To build an accurate hand-object detector for such unique application domains, a large amount of data and labels must be collected from scratch. However, data collection and annotation can be difficult due to various reasons, such as cost or privacy issues. In particular, expert knowledge is required to annotate the data in such specific application domains. Under these limitations, a hand-object detector may overfit the training data and lack the generalization ability due to the small amount of training data.

\begin{figure}[tb]
      \centering
      \begin{minipage}[t]{\linewidth}
        \centering
        \includegraphics[width=0.85\linewidth]{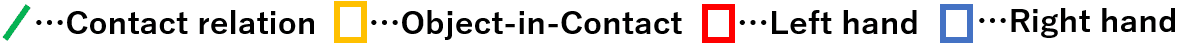}
      \end{minipage} \\

      \begin{minipage}[t]{0.48\linewidth}
        \centering
        \includegraphics[height=2.6cm, width=\linewidth]{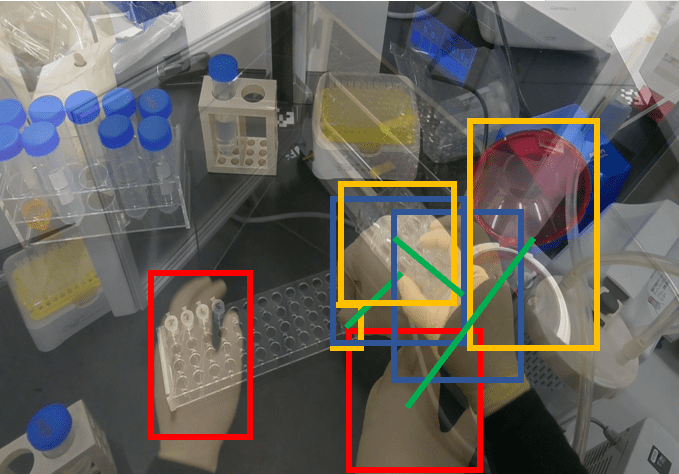}
        \subcaption{Ambiguous contact states}
      \end{minipage}
        \begin{minipage}[t]{0.48\linewidth}
            \centering
            \includegraphics[height=2.6cm, width=\linewidth]{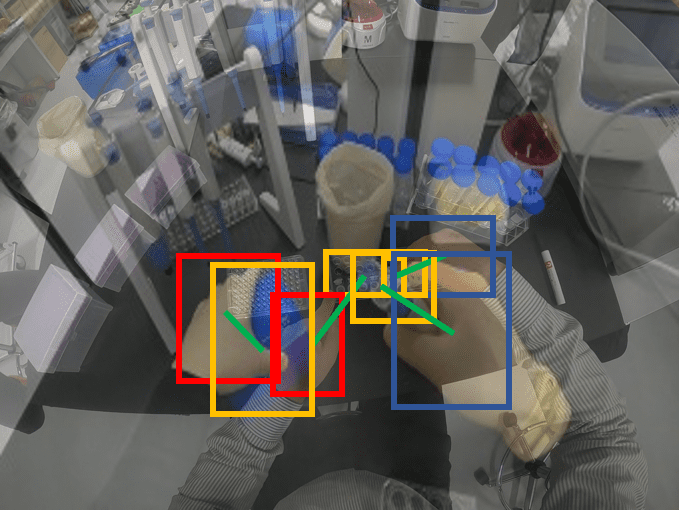}
            \subcaption{Ambiguous object boundaries}
        \end{minipage}
    \caption{\textbf{Problems with Mixup.}
    Naively mixing two images causes ambiguity in (a) contact states and (b) object boundaries.}
    \label{fig:mixup_issue}
\end{figure}

To improve the generalization ability of the detector trained on a small dataset, data augmentation is a key component in training.
Recently, Mixup~\cite{zhang2017mixup}, a method that overlays two different images, has been used as an empirically strong augmentation for object detection~\cite{zhou2021instant}.
Nevertheless, naively applying Mixup induces unintended biases in hand-object detection. As shown in Figure~\ref{fig:mixup_issue}, (a) contact states become ambiguous when hand-object pairs from different images overlap, and (b) the concentration of hands and objects in a specific local region makes identifying object boundaries difficult.
These unintended mixtures will degrade the performance of a hand-object detector.

\begin{figure}[tb]
  \centering
  \includegraphics[width=0.9\linewidth]{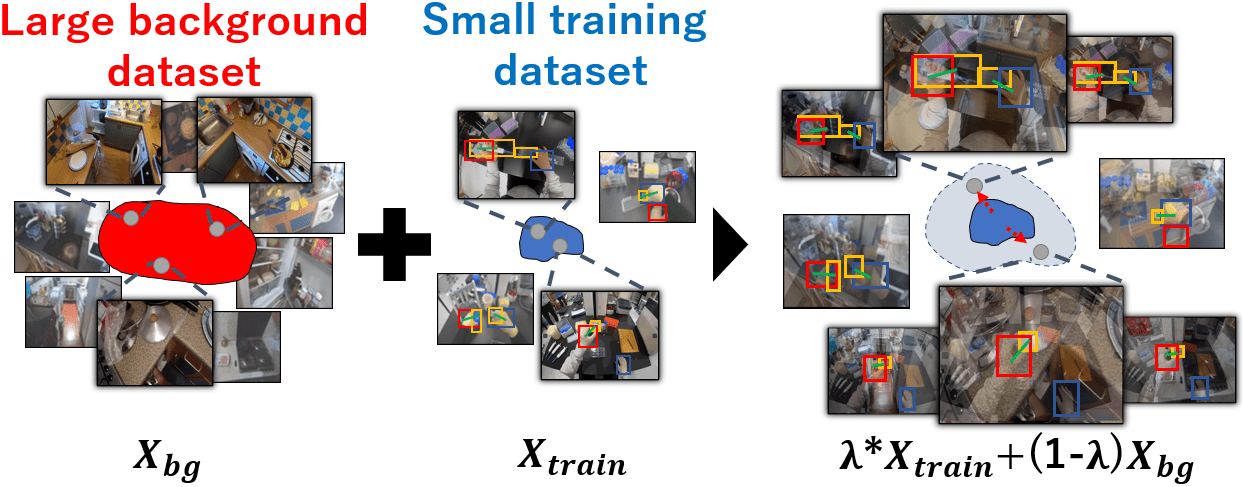}
  \caption{\textbf{Overview of Background Mixup.} We aim to improve diversity in training data while preserving foreground's semantics.}
  \label{fig:teaser}
\end{figure}

To handle this, we propose a novel data-augmentation method, called Background Mixup, that utilizes data-mixing regularization while reducing the unintended effects in hand-object detection. As shown in Figure~\ref{fig:teaser}, we aim to augment a training image by mixing it with the background of external image sources that does not contain the foreground (i.e., hands and objects-in-contact) and using the mixed images for training a hand-object detector. 
The contributions of this paper are summarized as follows.

\begin{itemize}
      \item We propose a novel data-augmentation method, Background Mixup, that mixes a training image and a background image to improve the generalization ability of a hand-object detector in a small dataset.

      \item Compared with Mixup, our experiments showed that Background Mixup improves the performance of a hand-object detector in supervised and semi-supervised learning settings.

      \item Our method has also shown to be effective in reducing the number of false-positive predictions although Mixup has the disadvantage in this metric.

\end{itemize}
\section{Related Work}
\label{sec:related}

\subsection{Hand and Object-in-Contact}
Jointly analyzing hands and objects-in-contact serves to understand human behavior~\cite{li2015delving,bambach2015lending}. While these studies have been conducted in a limited scale of data, Shan et al.~\cite{Shan20} proposed a large-scale dataset for training a hand-object detector localizing hands and interacting objects, which is collected in daily situations such as EPIC-KITCHENS 2018~\cite{damen2018scaling}, EGTEA~\cite{li2018eye}, CharadesEgo~\cite{sigurdsson2018actor}.
However, directly fine-tuning the hand-object detector on a small and specific dataset can lead to limited performance as discussed in Section~\ref{sec:intro}.

To overcome this, we developed Background Mixup to improve the generalization ability of a hand-object detector on a small dataset in specific domains such as biomedical experiments and factory work. 

\subsection{Mixture-Based Data Augmentation}
\label{sec:interpolation_aug}
Mixture-based data augmentation mixes input data with other inputs to increase the diversity of data on a small dataset and improve the generalization performance of the model.
Several mixture-based methods, such as Mixup~\cite{zhang2017mixup}, CutMix~\cite{yun2019cutmix}, Mosaic~\cite{ge2021yolox}, and Cutout~\cite{devries2017cutout}, have been used in many downstream tasks.

These mixture-based methods are used for semi-supervised learning of object detection~\cite{liu2021unbiased,zhou2021instant}. 
Unbiased-Teacher~\cite{liu2021unbiased} uses Cutout while Instant-Teaching~\cite{zhou2021instant} uses Mixup and Mosaic showing that Mixup particularly contributes to improving the performance of object detection.
However, applying Mixup leads to unintended biases in the hand-object detection, as discussed in Section~\ref{sec:intro}.
These unintended mixtures will degrade the performance of a hand-object detector.

\section{Proposed Method}
\label{sec:method}
In this section, we introduce our proposed training of a hand-object detector with  Background Mixup data augmentation.
Let $\mathcal{X}_{train}$ and $\mathcal{X}_{test}$ be sets of training and testing images, respectively.
When the size of $\mathcal{X}_{train}$ is small, a hand-object detector trained on $\mathcal{X}_{train}$ may not generalize well to $\mathcal{X}_{test}$ due to over-fitting to the training data.
To solve this problem, we propose Background Mixup that uses a background image without foreground entities, i.e., hands and objects-in-contact, for increasing the diversity of the training data.

We use a trained hand-object detector~\cite{Shan20} to extract the background images from an external image source (e.g., kitchens), which are different from our target data of $\mathcal{X}_{train}$ and $\mathcal{X}_{test}$.
We extract the background images in which neither object-in-contact nor hand was detected by the hand-object detector, and construct a set of the background images $\mathcal{X}_{bg}$.

\begin{figure}[tb]
  \centering
  \includegraphics[width=0.8\linewidth]{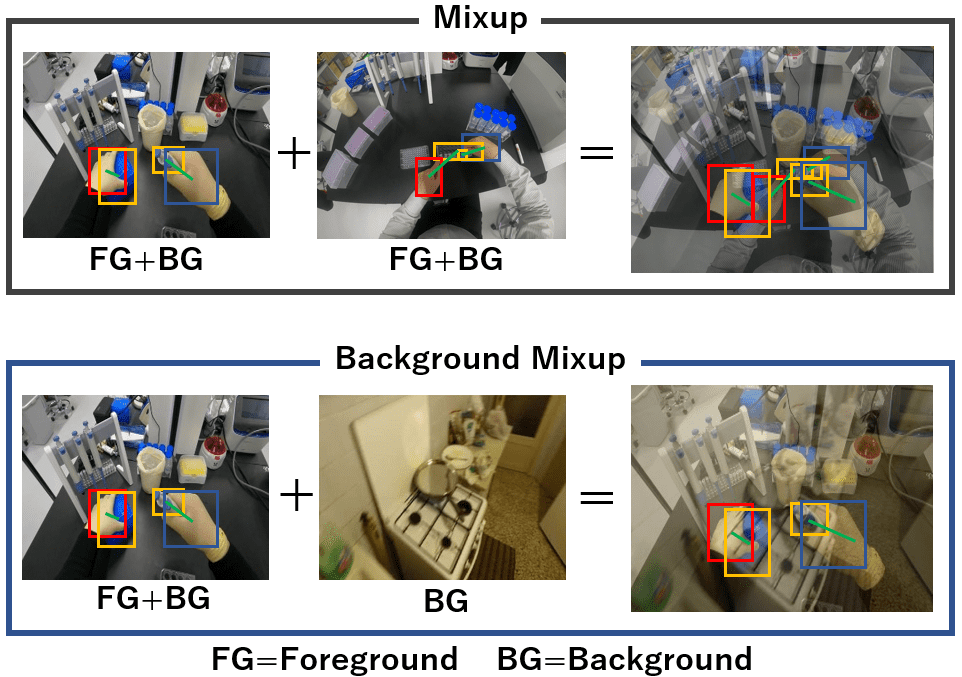}
  \caption{Comparison of Mixup~\cite{zhang2017mixup} and Background Mixup.}
  \label{fig:diff_mixup}
\end{figure}

Figure \ref{fig:diff_mixup} shows a comparison between Mixup and Background Mixup. 
With Mixup, the foreground and background are combined, causing unintended effects that make the contact state ambiguous or make it difficult to identify the boundaries of objects, as shown in Figure \ref{fig:mixup_issue}.
In contrast, Background Mixup reduces such unintended effects by mixing the training image with the background image, which can retain the foreground of the training image.

We denote the training image as $I_{train}\in\mathcal{X}_{train}$ and the background image as $I_{bg}\in\mathcal{X}_{bg}$. 
We define Background Mixup as:
\begin{equation}
  \hat{I} = \lambda I_{train} + (1-\lambda) I_{bg}
\end{equation}
\begin{equation}
  \lambda \sim Beta(\alpha, \beta).
\end{equation}

\noindent where $I_{train}$ and $I_{bg}$ are randomly sampled and $\lambda\in[0, 1]$ is a parameter controlling the degree of the mixture. Following Mixup, the parameter $\lambda$ is drawn from beta distribution $Beta(\alpha, \beta)$ where $\alpha$ and $\beta$ indicate hyperparameters to determine the distribution shape.

We use $\hat{I}$ for training the hand-object detector instead of $I_{train}$.
This method can be implemented with a small computational cost at the training stage. Thus no additional computational cost is required in inference.  
\section{Experiments}
\label{sec:experiment}
\subsection{Experimental Setup}

\textbf{Datasets.}
We validated our method on various hand manipulation datasets including biomedical experiments, mock factories, and kitchens.
We used a first-person video dataset of biomedical experiments and a mock factory environment dataset~\cite{ragusa2021meccano} as specific application domains where 
the data size and variety are limited. We also used a kitchen environment dataset~\cite{damen2018scaling, Shan20} including diverse cooking scenes.

For the dataset of biomedical experiments, we recorded 12 videos that contained basic actions such as preparing reagents in a biomedical lab. The bounding boxes of hands and objects-in-contact were annotated by an expert in the field. The total duration is 27 minutes, and the number of annotated frames is 3,093. We split the 12 videos into 6:3:3 for train:val:test.

For the set of background images in the experiments on biomedical and factory datasets, we used EPIC-KITCHENS-100~\cite{Damen2021RESCALING} of cooking scenes in kitchens as an external image source.
For the experiments on the cooking dataset, we used Something-Something V2~\cite{goyal2017something} of daily scenes as an external image source to augment the background appearance. 

\textbf{Training details.}
We measure the performance of our method in supervised and semi-supervised learning settings. For supervised learning, we fine-tune the pre-trained hand-object detector proposed by Shan et al.~\cite{Shan20}. For semi-supervised learning, we trained the hand-object detector in the training pipeline of Unbiased-Teacher (UB-Teacher)~\cite{liu2021unbiased}.
We evaluated the performance by the average precision (AP) of hands and objects-in-contact.
Note that we do not provide hand AP in an experiment with mock factory environment dataset~\cite{ragusa2021meccano} because the hand bounding boxes are not annotated.

\textbf{Baselines.}
We denote our proposed Background Mixup with EPIC-KITCHENS-100~\cite{Damen2021RESCALING} and Something-Something V2~\cite{goyal2017something} as \textbf{BG-Mix$_{K}$} and \textbf{BG-Mix$_{D}$}, respectively.
We prepared two variants of Mixup as comparison methods.
\textbf{Mixup} is the original Mixup that combines two different images within a dataset, and \textbf{Mixup$_{K}$} is Mixup that mixes a training image with a randomly selected image from EPIC-KITCHENS 2018~\cite{Shan20,damen2018scaling}.

\subsection{Quantitative Evaluation}

\subsubsection{Supervised Learning}
\label{sec:qual_sup}

\begin{table*}[t]
\begin{center}
\caption{Quantitative comparisons on supervised learning.}
\label{tab:sup_eval_ind_kitchen}

\scalebox{0.8}[0.8]{

    \begin{tabular}{@{}lcccccccc@{}}
    \toprule
         & \multicolumn{3}{c}{Biomedical} & \multicolumn{2}{c}{Factory} & \multicolumn{3}{c}{Cooking} \\ 
        Model & hand AP & obj AP & mAP & hand AP & obj AP & hand AP & obj AP & mAP \\ 
    
    \midrule
     Supervised  & \textbf{90.9} $\pm{0.0}$ & 70.6 $\pm{0.3}$ & 80.7 $\pm{0.2}$ & - & 45.0 $\pm{0.1}$ & \textbf{90.6} $\pm{0.0}$ & \textbf{66.4} $\pm{0.1}$ & \textbf{78.5}$\pm{0.0}$\\
     \; + Mixup & \textbf{90.9} $\pm{0.0}$ &  70.4 $\pm{0.1}$ & 80.6 $\pm{0.0}$ & - & 44.6 $\pm{0.0}$ & \textbf{90.6} $\pm{0.0}$ & 65.6 $\pm{0.2}$ & 78.1$\pm{0.1}$\\
     \; + Mixup$_{K}$ & \textbf{90.9} $\pm{0.0}$ & 69.8 $\pm{0.4}$ & 80.4 $\pm{0.2}$ & - & 44.6 $\pm{0.1}$ & - & - & - \\
     \; + BG-Mix$_{K}$ & \textbf{90.9} $\pm{0.1}$ & \textbf{72.2} $\pm{0.2}$ & \textbf{81.0} $\pm{0.1}$ & - & \textbf{45.2} $\pm{0.1}$ & - & - & - \\
     \; + BG-Mix$_{D}$ & - & - & - & - & - & \textbf{90.6} $\pm{0.0}$ & 65.7 $\pm{0.2}$ & 78.2$\pm{0.1}$\\
    \bottomrule
    \end{tabular}
}
\end{center}
\end{table*}

Table \ref{tab:sup_eval_ind_kitchen} lists the results on the supervised learning settings. 
In the biomedical and mock factory environment datasets, BG-Mix$_{K}$ exhibited the highest performance in mAP and object AP while Mixup and Mixup$_{K}$ decreased in these metrics. This is because our method avoids the unintended effects shown in Figure \ref{fig:mixup_issue}, which degrades the performance of detecting an object-in-contact.
In the cooking dataset, however, Mixup, Mixup$_{K}$ and BG-Mix$_{D}$ all obtained lower obj AP and hand AP than the Supervised baseline because the dataset already has diverse foreground and background appearances even without data augmentation.
The hand APs of Background Mixup did not increase because the hand APs were already saturated.

\subsubsection{Semi-Supervised Learning}

\begin{table*}[tb]
\begin{center}
\caption{Quantitative comparisons on semi-supervised learning at 1\% labels.}
\label{tab:semi_eval_ind_kitchen}

\scalebox{0.8}[0.8]{
    \begin{tabular}{@{}lcccccccc@{}}
    \toprule
         & \multicolumn{3}{c}{Biomedical} & \multicolumn{2}{c}{Factory} & \multicolumn{3}{c}{Cooking} \\ 
        Model  & hand AP & obj AP & mAP & hand AP & obj AP & hand AP & obj AP & mAP \\ 
    
    \midrule
     UB-Teacher & 90.6 $\pm{0.1}$ & 64.0 $\pm{1.1}$ & 77.3 $\pm{0.4}$  & - & 27.0 $\pm{0.5}$ &  \textbf{90.5} $\pm{0.0}$ & 41.4 $\pm{0.5}$ & 65.9 $\pm{0.3}$\\
     \; + Mixup & \textbf{90.9} $\pm{0.0}$ & 62.4 $\pm{3.4}$ & 76.6 $\pm{1.7}$ & - & 30.4 $\pm{1.4}$ & 90.4 $\pm{0.0}$ & 46.5 $\pm{0.1}$ & 68.5 $\pm{0.1}$\\
     \; + Mixup$_{K}$ & 90.8 $\pm{0.1}$ & 65.4 $\pm{0.6}$ & 78.1 $\pm{0.3}$ & - & 31.0 $\pm{1.4}$ & - & - & - \\
     \; + BG-Mix$_{K}$ & 90.8 $\pm{0.1}$ & \textbf{66.4} $\pm{0.1}$ & \textbf{78.6} $\pm{0.1}$ & - & \textbf{32.6} $\pm{0.7}$ & - & - & - \\
     \; + BG-Mix$_{D}$ & - & - & - & - & - & \textbf{90.5} $\pm{0.0}$ & \textbf{47.2} $\pm{0.5}$ & \textbf{68.9} $\pm{0.2}$\\
    \bottomrule
    \end{tabular}
    }
\end{center}
\end{table*}

Table \ref{tab:semi_eval_ind_kitchen} shows the results for semi-supervised learning with 1\% labeled data. BG-Mix$_{K}$ and BG-Mix$_{D}$ exhibited the highest performance on all datasets, except for the hand AP on the biomedical data.
In the cooking dataset having a variety of objects and backgrounds, although the performance of supervised learning decreased with both BG-Mix$_D$ and Mixup as shown in Section \ref{sec:qual_sup}, BG-Mix$_D$ improved the performance of semi-supervised learning.
This indicates that Background Mixup is effective under limited labels where the fully-supervised model suffers from generalizing to unknown test data.

\subsubsection{Analysis of False-positive Predictions}
Although mAP is a standard evaluation criterion in object detection, there is a technique that can improve the mAP score by allowing many false positives with low confidence ~\cite{kaggle2021map}. 
However, detection results that contain many false positives are problematic in real scenarios. 
Therefore, we experimented with precision to measure the percentage of false positives in detection results of a hand-object detector. Precision indicates the percentage of true positives among the predictions detected as positive. In other words, the lower precision, the higher the percentage of false positives.

Table \ref{tab:comparison_fp} shows a comparison of precision when the percentage of labeled data is 1\% in semi-supervised learning, and the confidence threshold is 0.1.
While Mixup$_{K}$ improves mAP, the precision is decreased.
This indicates training the detector on mixed images with many overlapping bounding boxes in a specific area, as illustrated in Fig.~\ref{fig:mixup_issue}(b), induces the bias of increasing the number of false positives.
BG-Mix$_{K}$ can improve the mAP without increasing the number of false positives because it keeps the information of hand-object contact and avoids the concentration of target hands and objects-in-contact in a local region.

\begin{table}[tb]
\begin{center}
\caption{Comparisons of false positive predictions on biomedical experiments dataset.}
\label{tab:comparison_fp}
    \scalebox{0.85}[0.85]{

    \begin{tabular}{@{}lccc@{}}
    \toprule
        Model & mAP & Precision$_{hand}$ & Precision$_{obj}$ \\ 
    \midrule
        UB-Teacher    & 77.3$\pm{0.6}$ & 87.1 $\pm{1.5}$  & \textbf{49.7} $\pm{0.1}$   \\
        \; +Mixup              & 76.6$\pm{1.7}$ & 76.8 $\pm{1.8}$  & 42.0 $\pm{3.2}$   \\
        \; +Mixup$_{K}$  & 78.1$\pm{0.3}$ & 75.6 $\pm{5.3}$  & 38.7 $\pm{2.9}$   \\
        \; +BG-Mix$_{K}$ & \textbf{78.6} $\pm{0.1}$ &\textbf{89.1} $\pm{2.5}$  & 48.8 $\pm{2.6}$  \\
    \bottomrule
    \end{tabular}
    }
\end{center}
\end{table}

\subsection{Qualitative Evaluation}

\begin{figure}[tb]
    \vspace{0.3cm}
    \hspace{-0.35cm}
    \begin{tabular}{c}
    \includegraphics[width=0.35\linewidth]{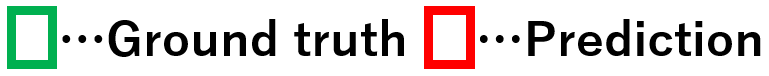} \\
    \begin{tabular}{cccc}
        \begin{minipage}[t]{0.23\linewidth}
            \centering
            \includegraphics[clip, width=\linewidth]{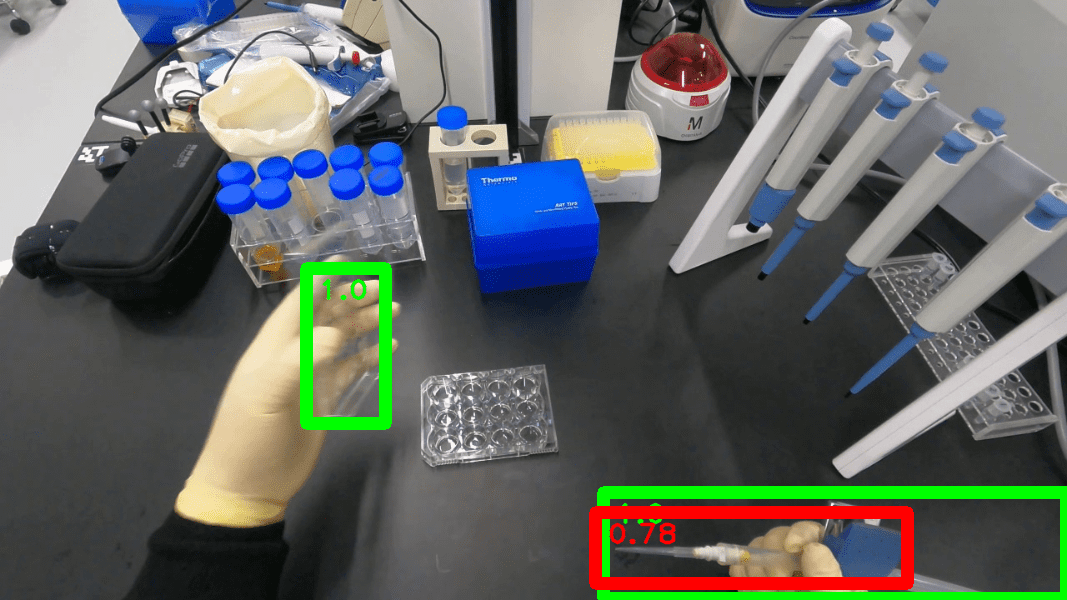} \\
            \includegraphics[clip, width=\linewidth]{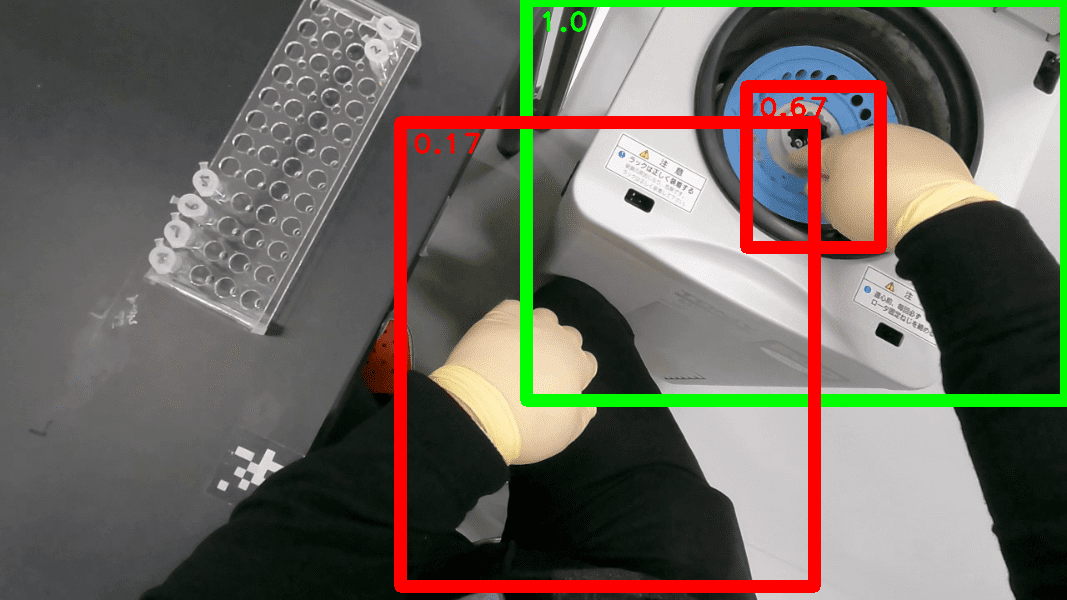} \\
            \includegraphics[clip, width=\linewidth]{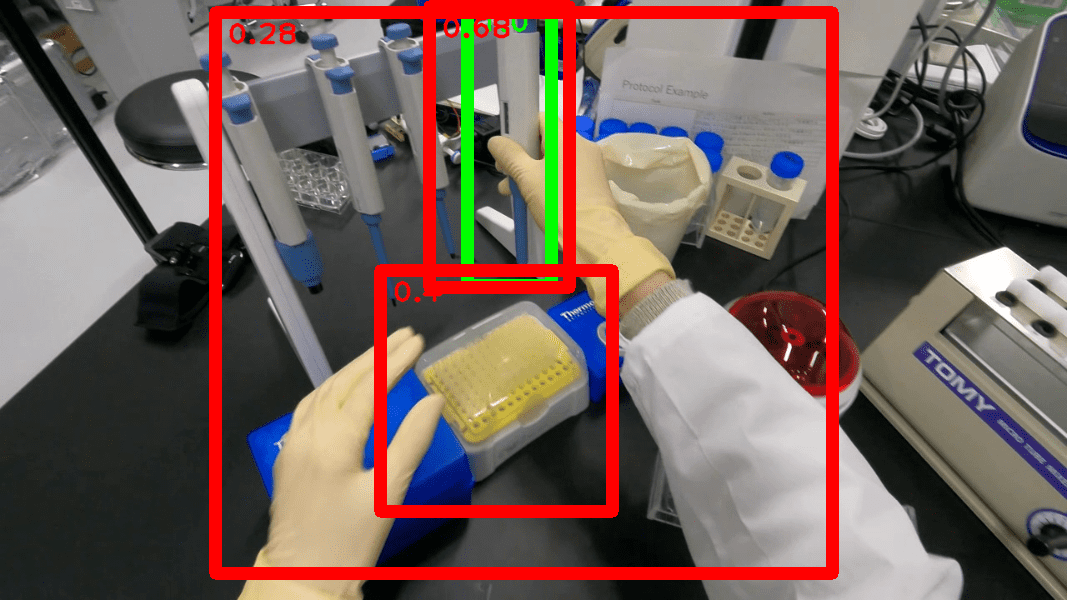} \\
            \includegraphics[clip, width=\linewidth]{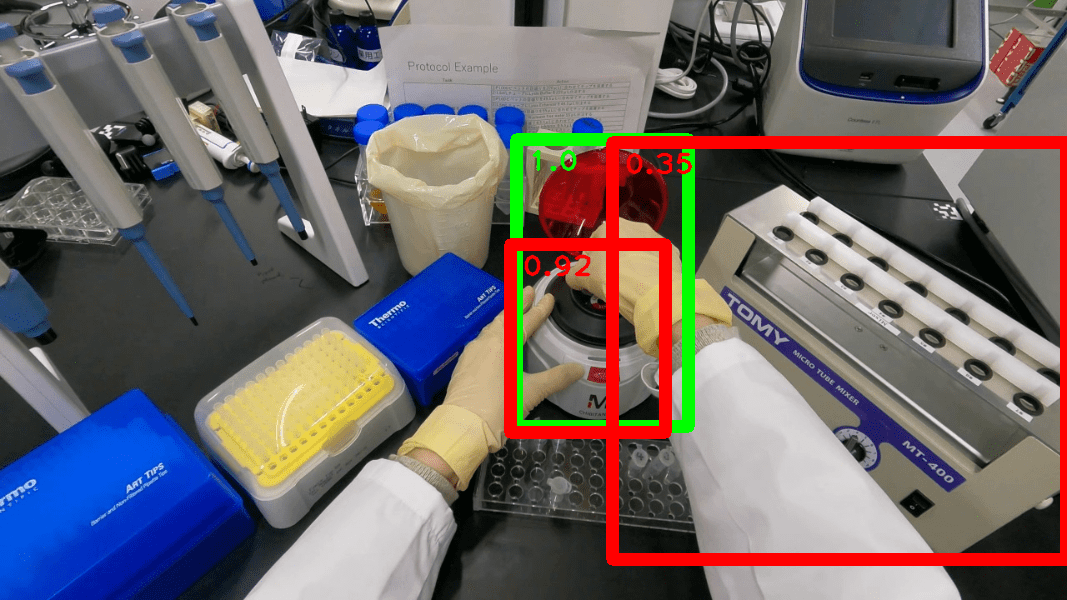} \\
            \subcaption{UB-Teacher}
        \end{minipage}
        
         \begin{minipage}[t]{0.23\linewidth}
            \centering
            \includegraphics[clip, width=\linewidth]{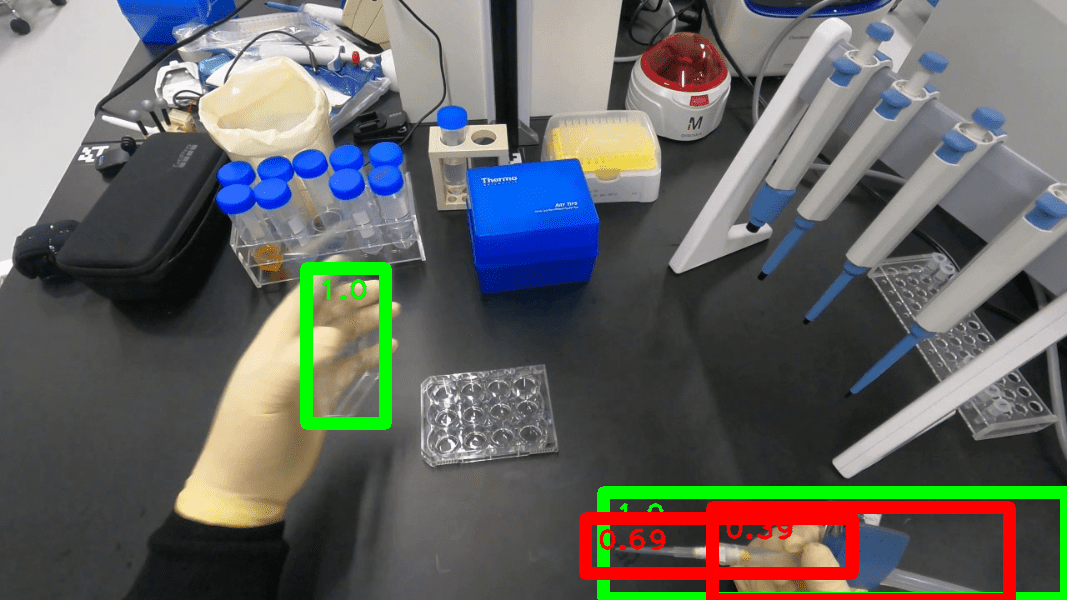} \\
            \includegraphics[clip, width=\linewidth]{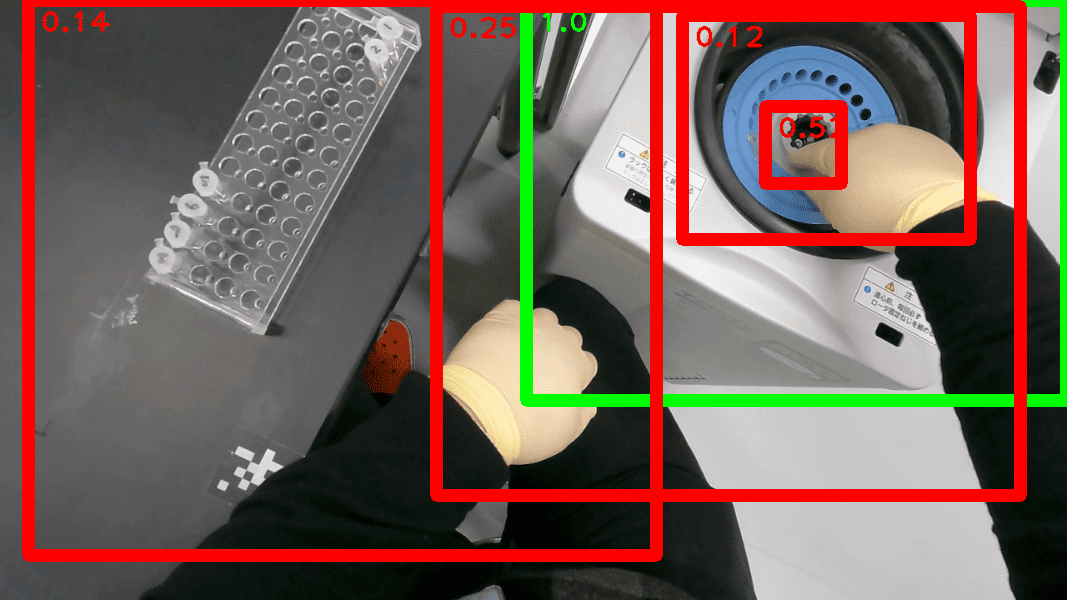} \\
            \includegraphics[clip, width=\linewidth]{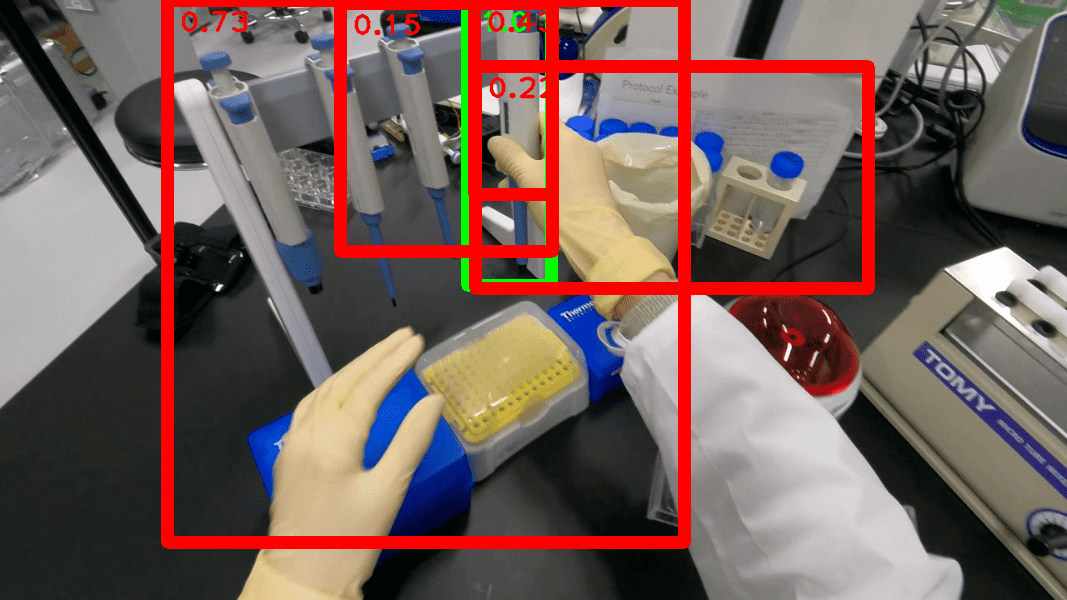} \\
            \includegraphics[clip, width=\linewidth]{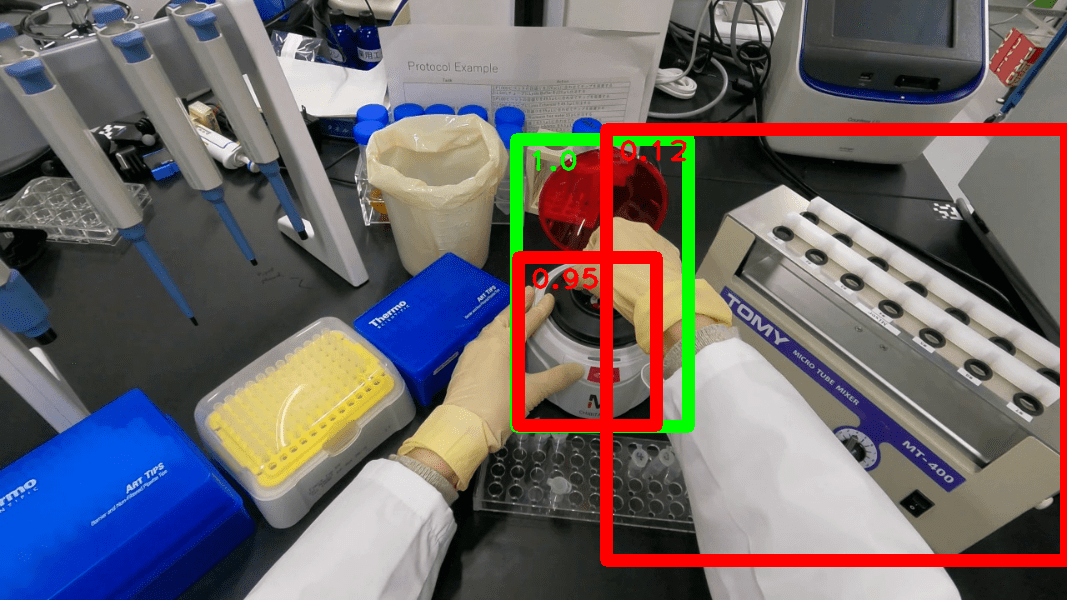} \\
            \subcaption{+Mixup}
         \end{minipage}
        
         \begin{minipage}[t]{0.23\linewidth}
            \centering
            \includegraphics[clip, width=\linewidth]{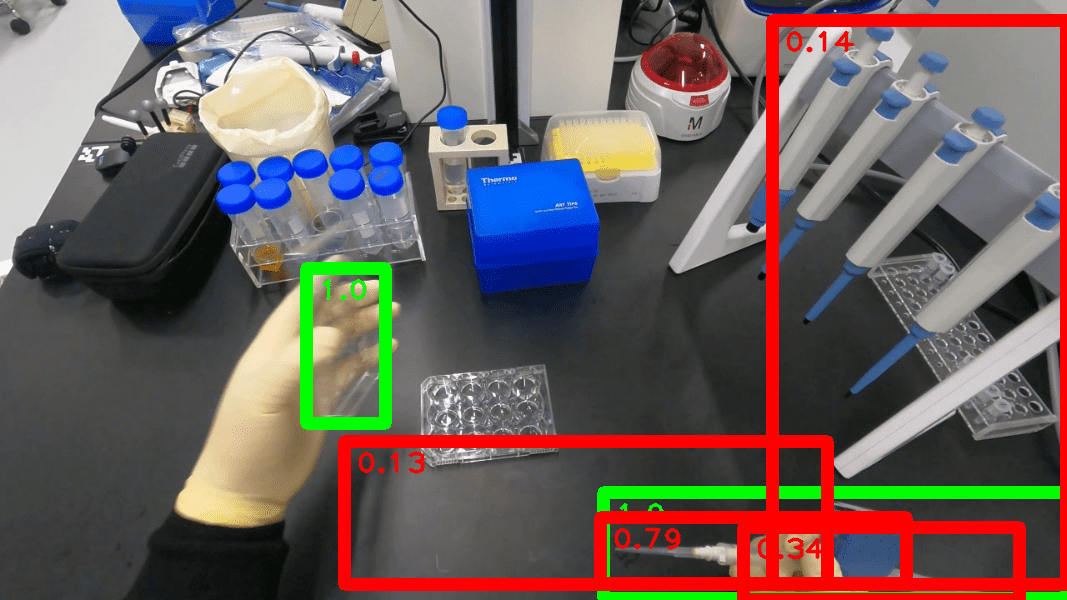} \\
            \includegraphics[clip, width=\linewidth]{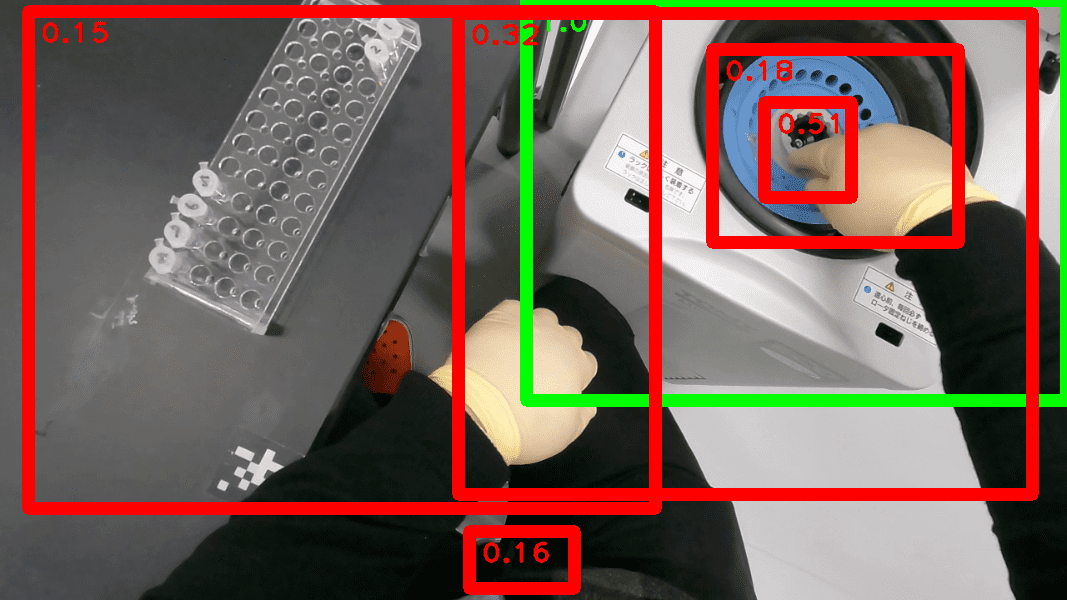} \\
            \includegraphics[clip, width=\linewidth]{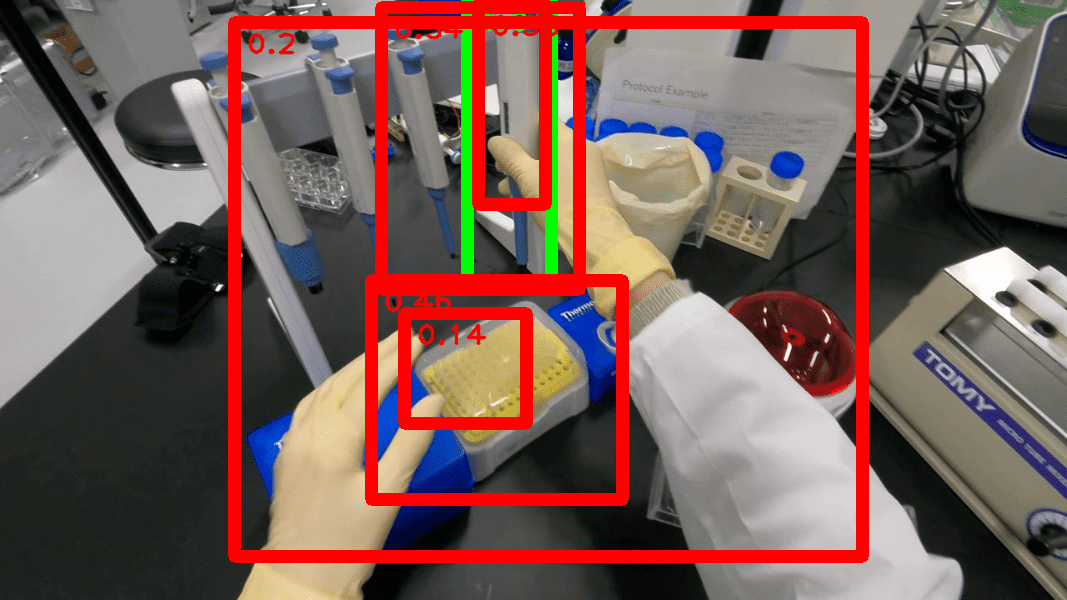} \\
            \includegraphics[clip, width=\linewidth]{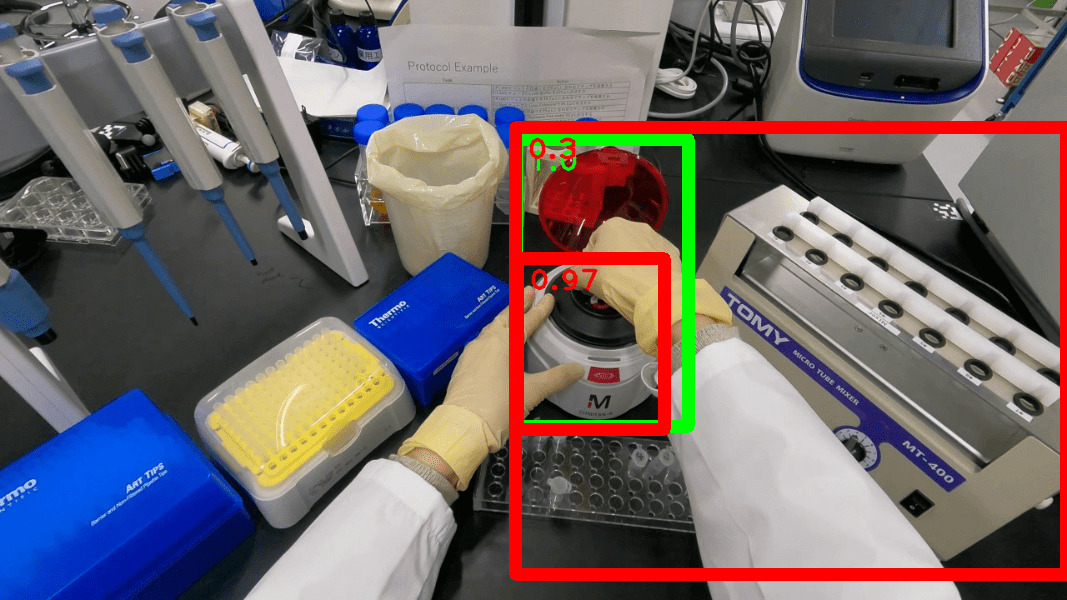} \\
            \subcaption{+Mixup$_{K}$}
         \end{minipage}
          
        \begin{minipage}[t]{0.23\linewidth}
            \centering
            \includegraphics[clip, width=\linewidth]{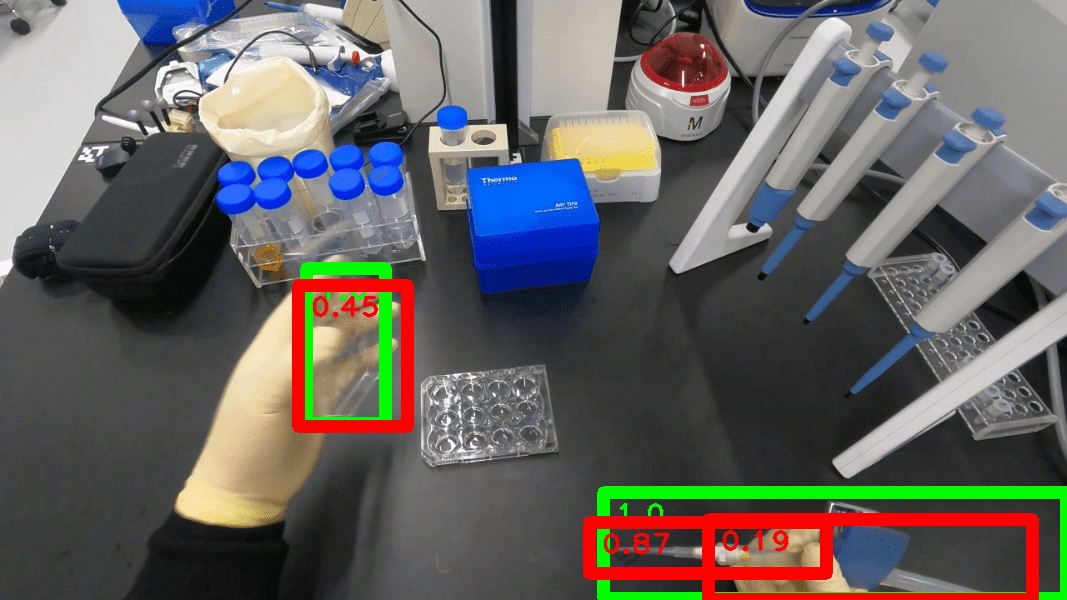} \\
            \includegraphics[clip, width=\linewidth]{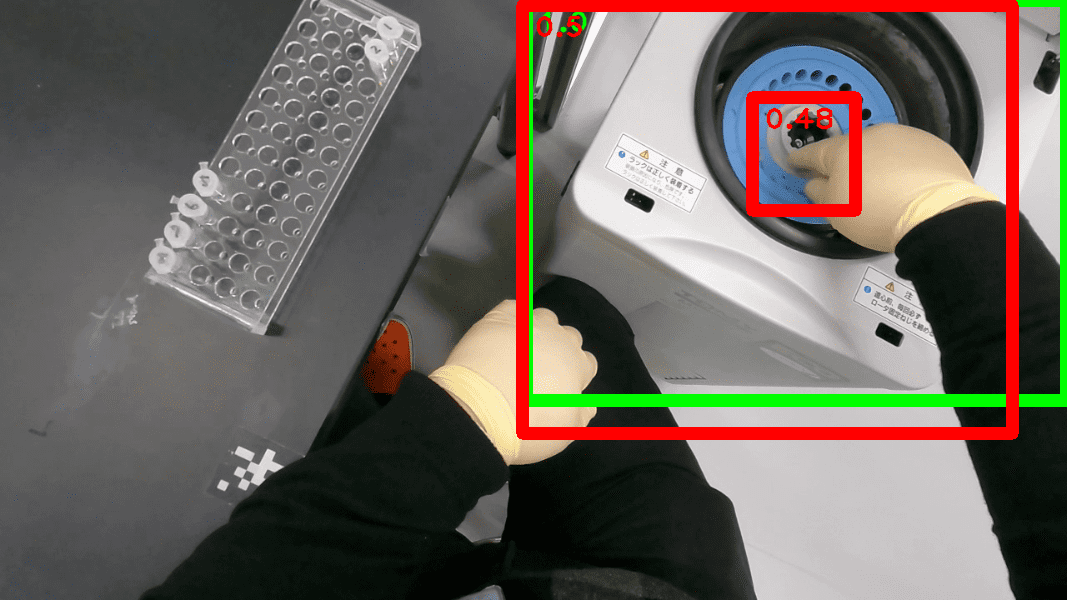} \\
            \includegraphics[clip, width=\linewidth]{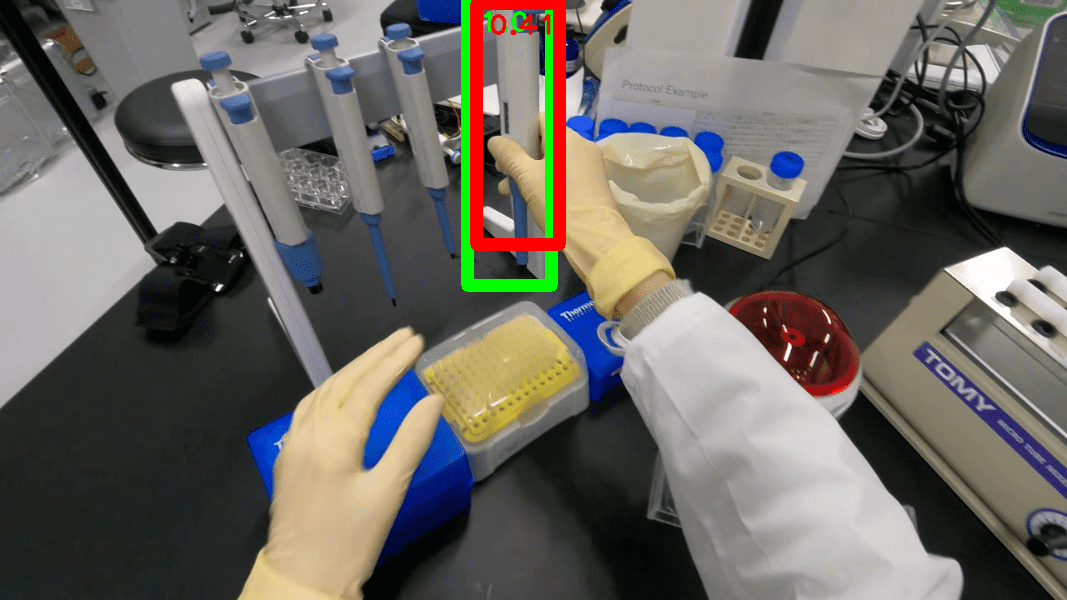} \\
            \includegraphics[clip, width=\linewidth]{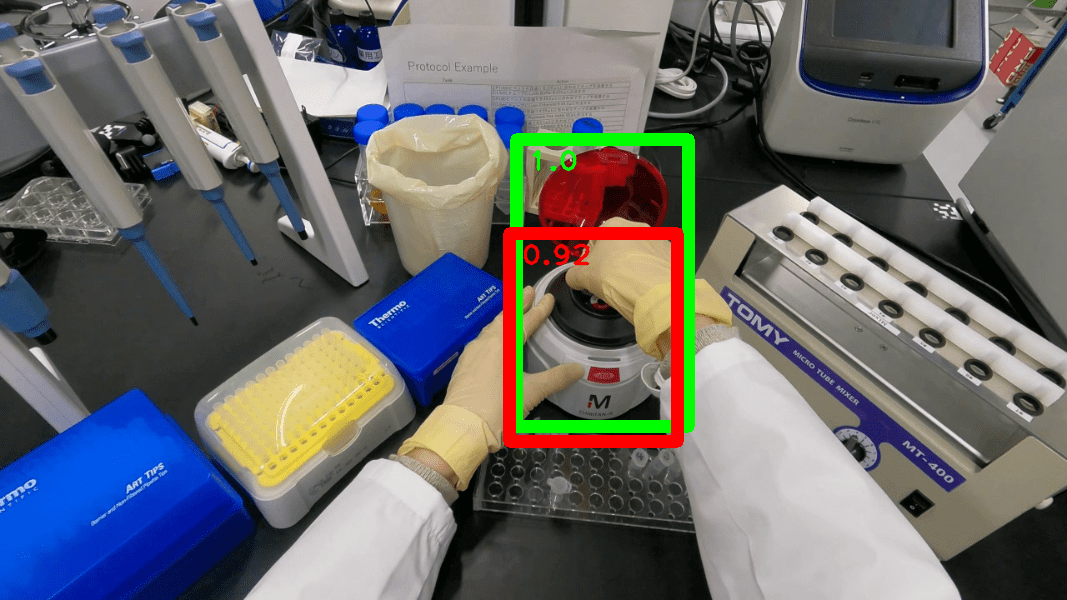} \\
            \subcaption{+BG-Mix$_{K}$}
        \end{minipage}

    \end{tabular}
    \end{tabular}

    \caption{Qualitative results in detecting object-in-contact. }
    \label{fig:qualitative_comparison}
\end{figure}

Figure \ref{fig:qualitative_comparison} shows the inference results for object-in-contact with a confidence threshold of 0.1. 
We observed that Figure~\ref{fig:qualitative_comparison} (b) Mixup and (c) Mixup$_{K}$ increased the number of false positives (e.g., red bounding boxes far from the ground truth bounding boxes).
In contrast, the predictions of Figure~\ref{fig:qualitative_comparison} (d) Background Mixup is less noisy and accurately represent the location of the object-in-contact compared to ground truth.
Our method of increasing the diversity of the background without changing the foreground semantics could improve the performance of a hand-object detector without increasing the number of false positives.

\section{Conclusion}
\label{sec:conclusion}

We proposed Background Mixup, which mixes training images with background images that do not contain the hands and the objects-in-contact, whereas Mixup mixes both the foreground (i.e., the hand and the object-in-contact) and the background. 
Background Mixup can improve the performance of a hand-object detector in small datasets, such as biomedical experiments and mock factory environments, by increasing the diversity of the background appearances while inhibiting the unintended effects caused by Mixup.
We have also shown that Background Mixup was effective in reducing the number of false positives.

\bibliographystyle{IEEEbib}
\bibliography{refs}

\clearpage

\end{document}